\documentclass[11pt,a4paper]{article}
\usepackage[hyperref]{acl2020}
\usepackage{times}
\usepackage{latexsym}

\usepackage{graphicx}
\usepackage{subfigure}
\usepackage{hyperref}

\usepackage{enumitem}
\setlist[itemize]{noitemsep, topsep=2pt}
\usepackage{microtype}

\aclfinalcopy 

\title{TypeShift: A User Interface for Visualizing the Typing Production Process}

\author{Adam Goodkind \\
  Northwestern University \\
  \texttt{a.goodkind@u.northwestern.edu} \\}

\date{\today}

\begin{document}
\maketitle

\begin{abstract}
TypeShift is a tool for visualizing linguistic patterns in the timing of typing production. Language production is a complex process which draws on linguistic, cognitive and motor skills. By visualizing holistic trends in the typing process, TypeShift aims to elucidate the often noisy information signals that are used to represent typing patterns, both at the word-level and character-level. It accomplishes this by enabling a researcher to compare and contrast specific linguistic phenomena, and compare an individual typing session to multiple group averages. Finally, although TypeShift was originally designed for typing data, it can easy be adapted to accommodate speech data, as well.\footnote{A web demo is available at \url{https://angoodkind.shinyapps.io/TypeShift/}. The source code can be accessed at \url{https://github.com/angoodkind/TypeShift}.} 
\end{abstract}

\section{Introduction}


The task of ``visualizing language production'' is both broad and difficult to implement conclusively. Common visualizations relating to language production include dendrograms, word clouds and frequency counts. These summary visualizations, however, only provide a static snapshot of language. They do not capture the dynamics elements that go into the process of language production, replete with its hesitations, inconsistencies and repairs.

Capturing the dynamics of language production is important because information is transmitted in language not only via the words chosen \citep[e.g.][]{chenoweth2001fluency}, but also by tone of voice, pauses and slips of the tongue \citep[e.g.][]{bosker2013makes,goldrick2016automatic}. Therefore, in order to accurately study language production, a researcher not only needs a record of the words that were produced, but also a record of \emph{how} they were produced.

The TypeShift tool aims to visualize the dynamics of language produced via a computer keyboard. Compared to speech production data, keyboard typing production data is relatively easy to capture \citep{priva2010constructing}. A simple keylogger can record when a key is depressed and then released, making it relatively easy to measure when a word begins and when it ends. Similar timing metrics are much more laborious and ambiguous with speech production data, as it is more difficult to determine the boundaries of phonemes in a speech stream. Nonetheless, the TypeShift tool can be readily adapted to speech data, if sufficient data is collected, requiring only minor alterations to the tool.

\subsection{Overall Motivation}

The importance of a system such as TypeShift lies in answering the multifaceted question, ``How can displaying temporal and dynamic information about language production elucidate linguistic processing?'' This question is broad and difficult to answer concisely. However, certain aspects of a typing session can shed light on the question, and provide partial answers. Below is a list of examples, with elements of a typing session that TypeShift can help to shed light on:

\begin{itemize}
\itemsep0em 

\item Is this particular typing instance fast or slow, both as compared to other typing instances from the same user, as well as compared to different subsets of the general population?

\item Does a typist produce certain linguistic structures, e.g. noun phrases or function words, in a fashion distinct from other linguistic structures, e.g. verb phrases or content words? Does a certain structure result in slower production or more revisions?

\item Does this typist produce language at a consistent rate, or do certain linguistic elements result in heavy revisions, or bursts of quick typing interspersed with slow typing?
\end{itemize}

Answers to these questions can be useful for investigating psycholinguistic questions, as well as for tasks in related areas such as machine learning. For example, typing patterns have been shown to be a proxy for shallow syntactic parsing \citep{plank2016keystroke}. However, in creating a training set, it is important to know if the typing data being used is typical or atypical, both for that typist and compared to others.

Finally, by being able to compare different trends, as well as investigate only certain linguistic elements and see the details of just one token, the TypeShift tool permits for targeted comparisons that are revealed through details-on-demand, an important principle of data visualization \citep{shneiderman1996eyes}. Using this capability, a cognitive scientist could use the tool to visualize how different linguistic structures affect processing, vis-a-vis pauses and production rates.

We begin by explaining the value of capturing keystroke information, and then outline how TypeShift builds upon and improves over previous methods of visualizing the typing process. Next, we briefly describe the large dataset around which the tool was originally built (although it is now easily extensible). We then describe the system design and its research motivations, using elucidating examples.

\section{Related Work}

\subsection{Why keystrokes?}

Keystroke dynamics is the science of capturing the detailed timing information of a typist's keystrokes \citep{moskovitch2009identity}. This timing information can vary from a parameter as coarse as the overall mean typing rate to a metric as fine-grained as the pause time between two particular keys. The details of a typing session can help predict many personal properties, from overall identity authentication \citep[][\textit{inter alia}]{ali2015authentication}, to emotion and cognition prediction \citep{epp2011identifying,brizan2015utilizing}. Further, it can convey social information and influence impression-formation in dyads and groups \citep[e.g.][]{kalman2013online}.

Pausing within language production can also be especially informative. A pause can signal everything from increased cognitive effort, due to performing lexical retrieval \citep{erman2007cognitive}, to a more intense focusing on the task at hand \citep{schilperoord1996s}.

\subsection{Previous visualization work}

\begin{figure}[t]
    \centering
    \frame{\includegraphics[width=\columnwidth]{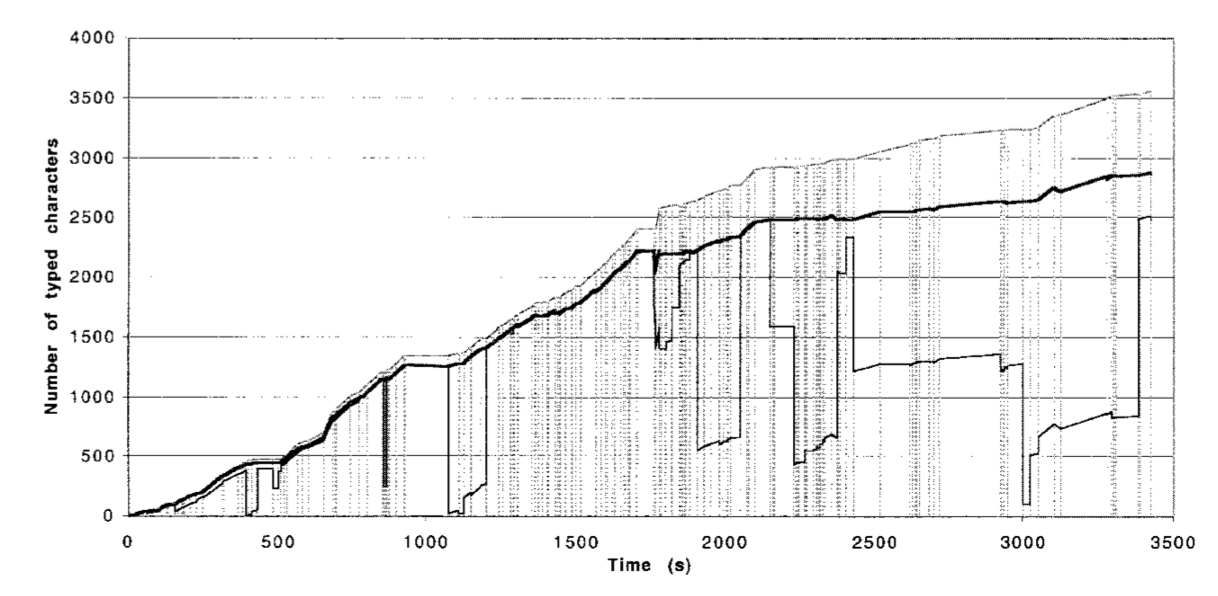}}
    \caption{An example of an LS Graph, a linear representation of the typing production process. The lighter top line represents the total number of keystrokes produced. The black line in the middle represents the number of characters appearing on the page, i.e. the top line minus revisions. The lowest line represents where the user is working at that time. Finally, the vertical lines represent where revisions have occurred.}
    \label{fig:lsgraph}
\end{figure}

Given that the dynamics of typing can provide such a large amount of information, many researchers have devised methods to visualize a typing session. For a recent overview, see \citet{becotte2017writing}. Many of the visualization techniques proposed for the typing process are limited to the ``what'' of typing, i.e. which keys were pressed and in what order. Some visualizations do focus on the rate of typing, which is reflected as a line chart where a steeper slope reflects a more rapid typing rate. This is illustrated in Fig. \ref{fig:lsgraph}, which is an example of an LS Graph \citep{lindgren2002ls}. However, this visualization only provides overall continuous lines, and lacks a discrete representation of separate words.

\begin{figure}[ht]
    \centering
    \frame{\includegraphics[width=.5\columnwidth]{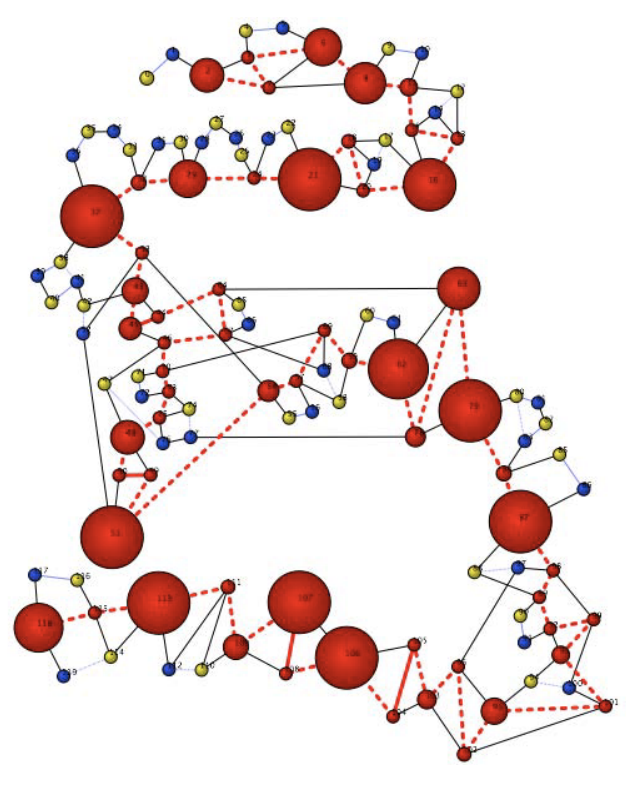}}
    \caption{Graph representation of the online writing process}
    \label{fig:graphnode}
\end{figure}

One visualization technique that \textit{does} represent word tokens can be seen in Fig. \ref{fig:graphnode} \citep{caporossi2011online}. While the discrete nodes show word tokens as well as the relationship between the order in which tokens were produced, this visualization fails to capture the temporal dynamics or varying rates of the LS Graph. In fact, the \emph{x,y} coordinates of this graph are completely arbitrary, and are arranged solely to present a compact visualization.

The visualization techniques described in this paper attempt to synthesize the best aspects of the visualization techniques mentioned above. It captures both the continuous, dynamics rate of language production, and highlights the discrete nature of word tokens.

\section{A New Keystroke Dataset}
This system was originally designed for use with a large, open-source dataset of typing data.\footnote{The data is available at \url{http://www2.latech.edu/~mike/TypingForTenData}} Because of the scale and complexity of the data collected, the dataset can be utilized to answer a number of research questions, on its own. Nonetheless, though, the flexibility of the TypeShift system allows researchers to import their own keystroke or speech data, so long as certain parameters are included.

The typing data was collected from 1,013 university students, seated in front of a desktop computer. Each participant typed responses to 10--12 question prompts, and each answer had to be at least 300 characters. This resulted in approximately 1 million words consisting of 10 million keystrokes. Participants included all genders, were both left- and right-handed, were both native and non-native English speakers, and reported a wide variety of school majors and typing/computer experience. Further details are provided in \citet{locklear2014continuous} and \citet{brizan2015utilizing}.

Each prompt was created in order to elicit one of six cognitive tasks: \textit{Remember}, \textit{Understand}, \textit{Apply}, \textit{Analyze}, \textit{Evaluate} or \textit{Create}. This task label was determined by the experimenters, who assessed the cognitive demands of each question as they related to Bloom's Taxonomy of Learning \citep{krathwohl2009taxonomy}. Examples prompts include:

\begin{itemize}
    \item (Remember) \textit{List the recent movies you've seen or books you've read.}
    \item (Evaluate) \textit{Do you think it's a good idea to raise tuition for students in order to have money to make improvements to the University? Why or why not?}
\end{itemize}

\section{Design Motivations}

\subsection{Data Preprocessing}

The collection procedures above resulted in an extensive amount of data. In order to understand the data with which we were dealing, the TypeShift tool was designed with the intention of creating meaningful data partitions, to make keystroke streams understandable and usable. To better illustrate the design decisions, we present a number of motivating examples below.

In order to first process the linguistic data, the stream of typing was clustered into word tokens using the Stanford CoreNLP parser \citep{manning2014stanford}. Each word token was annotated with its part-of-speech and semantic role. The semantic role is of interest because some psycholinguists posit that function words, e.g. \emph{you, the, it} or \emph{them}, are particularly psychologically informative \citep{chung2007psychological}. This is in contrast to the traditional view that content words, e.g. \emph{red, fast, smack} or \emph{mug}, contain the most amount of cognitive information. In addition, the start time and end time of each token was recorded, along with the keystrokes that comprise each token.

Typing sessions were not of uniform length, which could make comparisons difficult. To allow for typing sessions of different time lengths to be compared, the elapsed time of each typing session was normalized and converted to the proportion of a complete session, i.e. 0.0--1.0. Similarly, the typing rate for each token (\# of keystrokes/token time span) was normalized and converted to a \emph{z}-score, to allow for comparison across typists. By normalizing the typing rate, a visualization can shed light on not only whether a typist is typing rapidly \emph{overall}, but also rapidly \emph{for himself or herself}.

\begin{figure*}
    \centering
    \frame{\includegraphics[width=.7\textwidth]{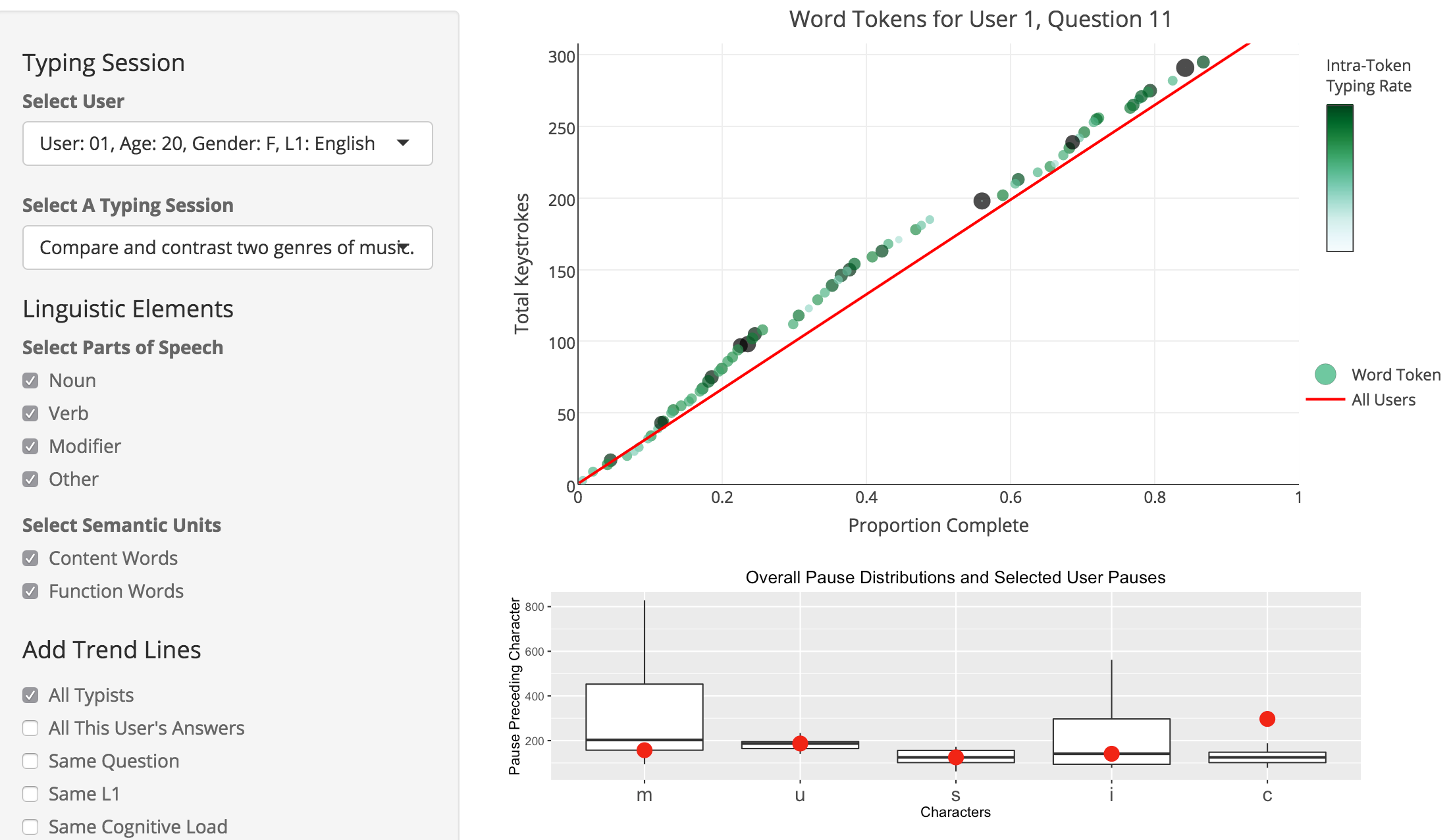}}
    \caption{Screenshot of the TypeShift User Interface}
    \label{fig:typeshift}
\end{figure*}

\subsection{TypeShift User Interface}

The TypeShift tool was designed using the Shiny Web Application Interface for \texttt{R} \citep{chang2016shiny}. The tool allows a user to select exactly which typing session to visualize, as well as which aspects of a typing session to include in the visualization.

A screenshot of the interface is seen in Fig. \ref{fig:typeshift}. The menu on the side provides a number of options:

\begin{itemize}
\itemsep0em 

\item
  {Selected User} -- A list of available typists is presented along with the the typist's age, gender and native language (L1).
\item
  {Typing Session} -- Within the selected typist, a list of the question prompts he or she responded to is presented.
\item
  {Parts-of-Speech} -- The user can choose to display only certain syntactic categories, such as only nouns or only verbs.
\item
  {Semantic Units} -- The user can choose to display only function or content words.
\item
  {Trend Lines} -- The user can compare this typing session to a variety of overall trends:

  \begin{itemize}
  \item
    {All Typists} -- The entire dataset
  \item
    {All This User's Answers} -- Every answer from the selected typist
  \item
    {Same Question} -- All responses to the same typing prompt
  \item
    {Same L1} -- All typing sessions from users with the same native language
  \item
    {Same Cognitive Load} -- All responses to typing prompts that require a similar level of cognitive effort
  \end{itemize}
\end{itemize}

The main plot charts the total number of keystrokes produced versus the proportion of the typing session that was completed. In this way, the slope of the resulting line (series of points) reflects the typing rate: a steeper slope represents a more rapid typing rate. The color and size of a point represent the typing rate of a particular token, in contrast to the overall typing rate. For example, if a user types a single word faster than the other words, then the point representing that word will be larger and darker than the other points. Finally, a point's border thickness represents the number of times a token was revised. In other words, a token with a thick border represents a word produced with a high number of backspaces or deletes.

If a user wishes to investigate the character-by-character specifics of a particular token, she can click on any individual word, to see a detailed view of that token's timing (lower-right portion of Fig. \ref{fig:typeshift}). The pause time before each character is displayed as a red point for that particular instance. In addition, a box plot for the overall population distribution of pauses before that character is also displayed. In this way, a user can ascertain how a typist produces individual characters of a specific word, relative to other typist's producing the same word.

\subsection{Design Choice Motivations}

The design motivations of the TypeShift system are best understood through examples. The examples below highlight how TypeShift can bring to light key attributes of a typist or typing session.

\begin{figure}
    \centering
    \frame{\includegraphics[width=\columnwidth]{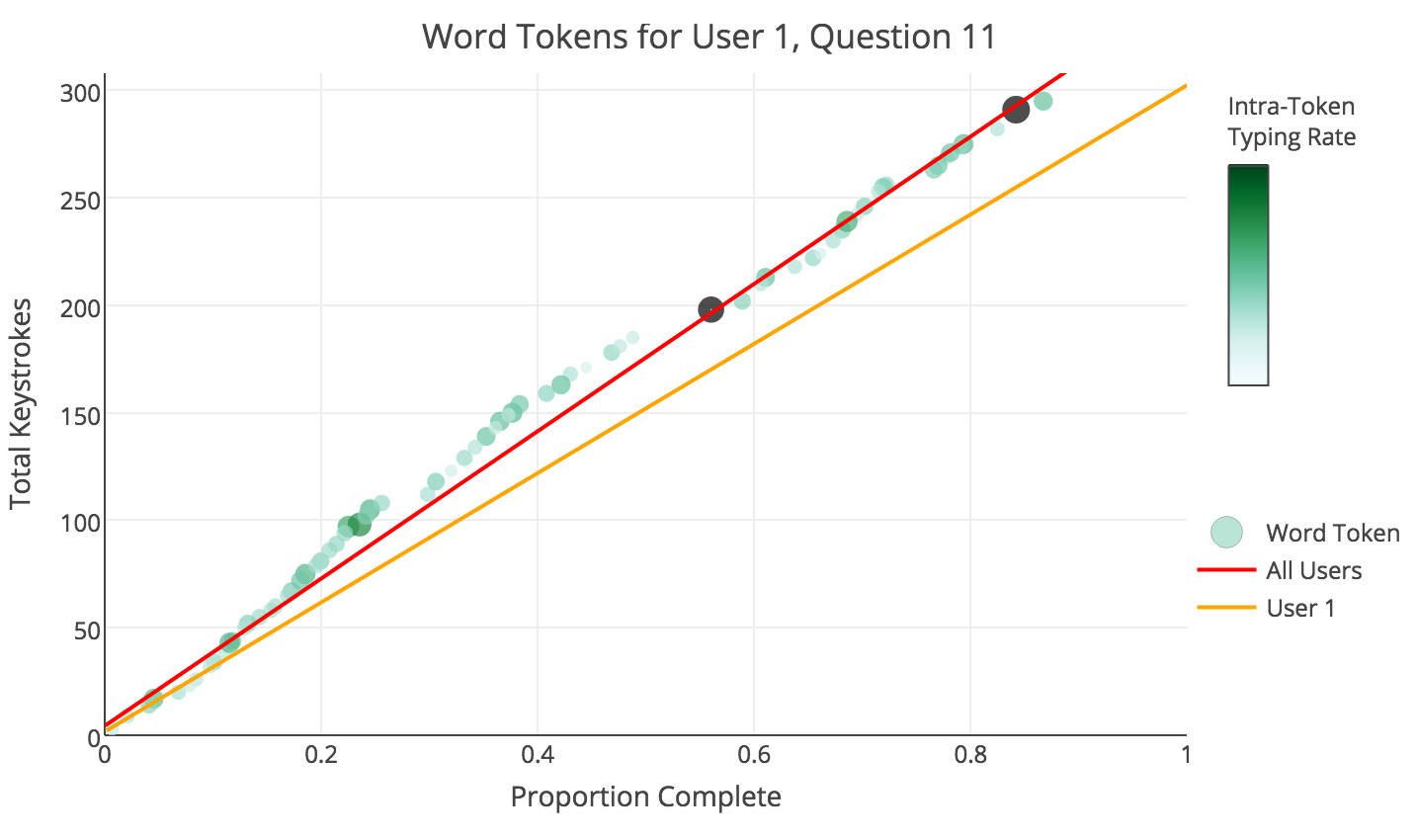}}
    \caption{Average typing session for a slow typist}
    \label{fig:avgtyping}
\end{figure}

Fig. \ref{fig:avgtyping} illustrates the importance of comparing a typing session not only to the overall, across-subjects mean, but to the within-subject mean, as well. Compared to the overall population (red line), this typing session would be considered average. However, for this typist, this particular session is actually very rapid, as the timing is above the user mean (orange line). If a researcher were trying to understand this particular typist, this visualization helps show that this is not a typical typing session for this typist.

\begin{figure}
    \centering
    \frame{\includegraphics[width=.7\columnwidth]{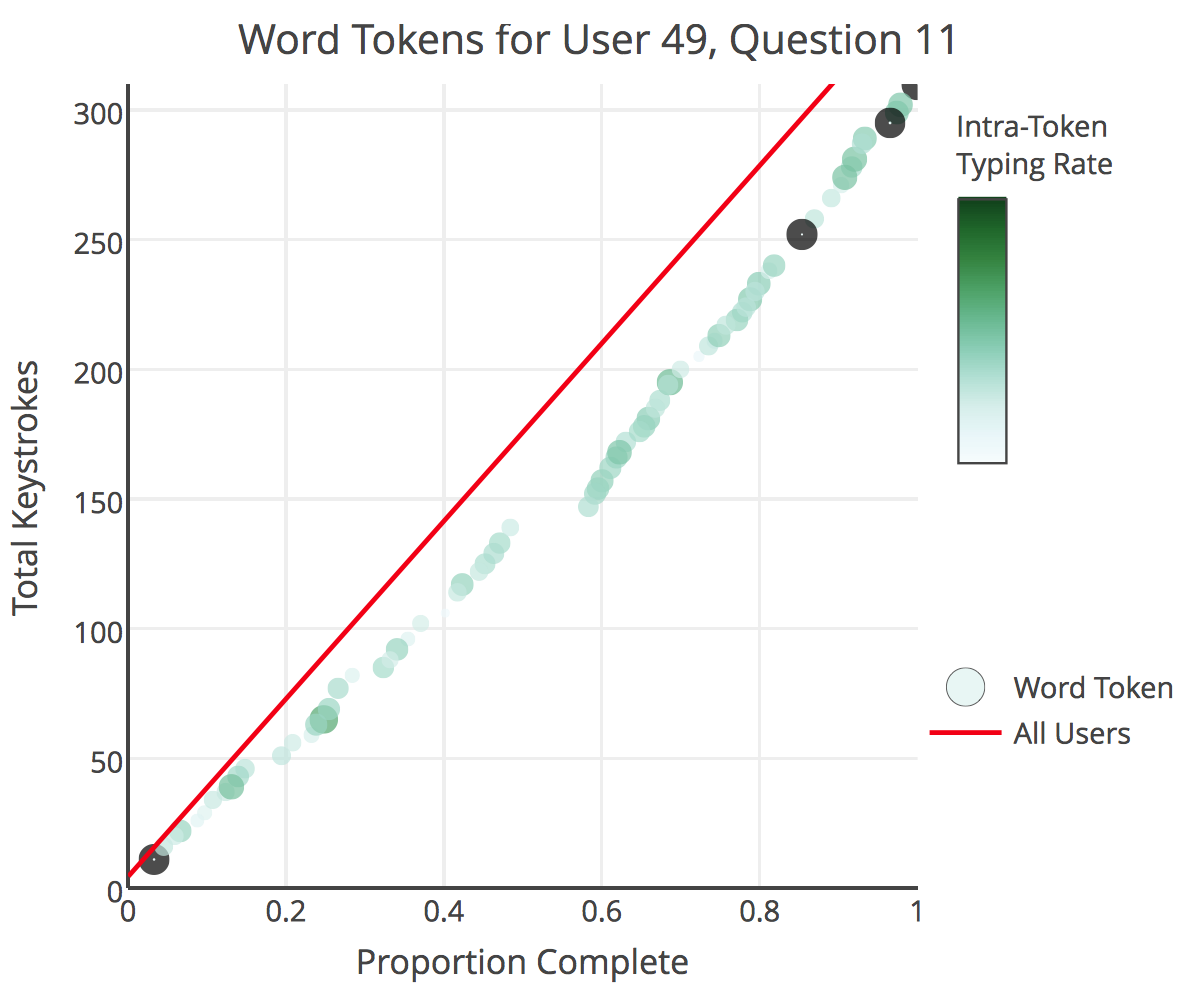}}
    \caption{Slow but steadily-paced typist, with typing bursts}
    \label{fig:slowtypist}
\end{figure}

\begin{figure}
    \centering
    \frame{\includegraphics[width=.7\columnwidth]{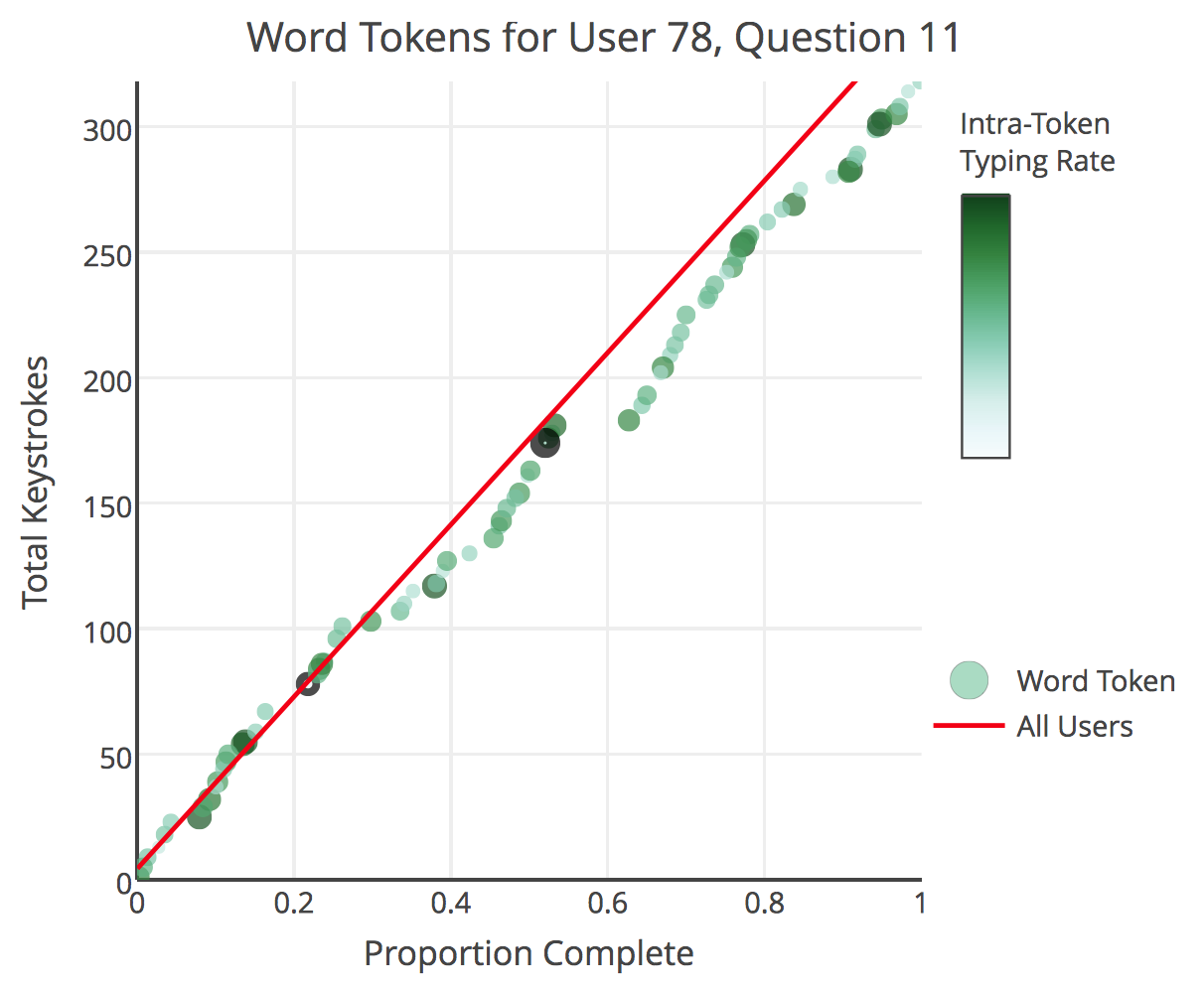}}
    \caption{Inconsistent typing rate}
    \label{fig:inconsistenttypist}
\end{figure}

Fig. \ref{fig:slowtypist} illustrates a typist who is slow, but types in bursts at a relatively constant rate. The overall constant rate is illustrated by the uniform slope of the plot, along with similarly sized and colored points. In contrast, Fig. \ref{fig:inconsistenttypist} shows a more inconsistent typing rate, with varying slopes and different colored points. The burstiness of the typing session is illustrated by the large gaps in the plot, when typing pauses for a significant period of time. By being able to select only certain linguistic structures, or zoom in on a single word, a researcher could investigate exactly what is \textit{causing} these typing bursts.

\begin{figure}
    \centering
    \frame{\includegraphics[width=.7\columnwidth]{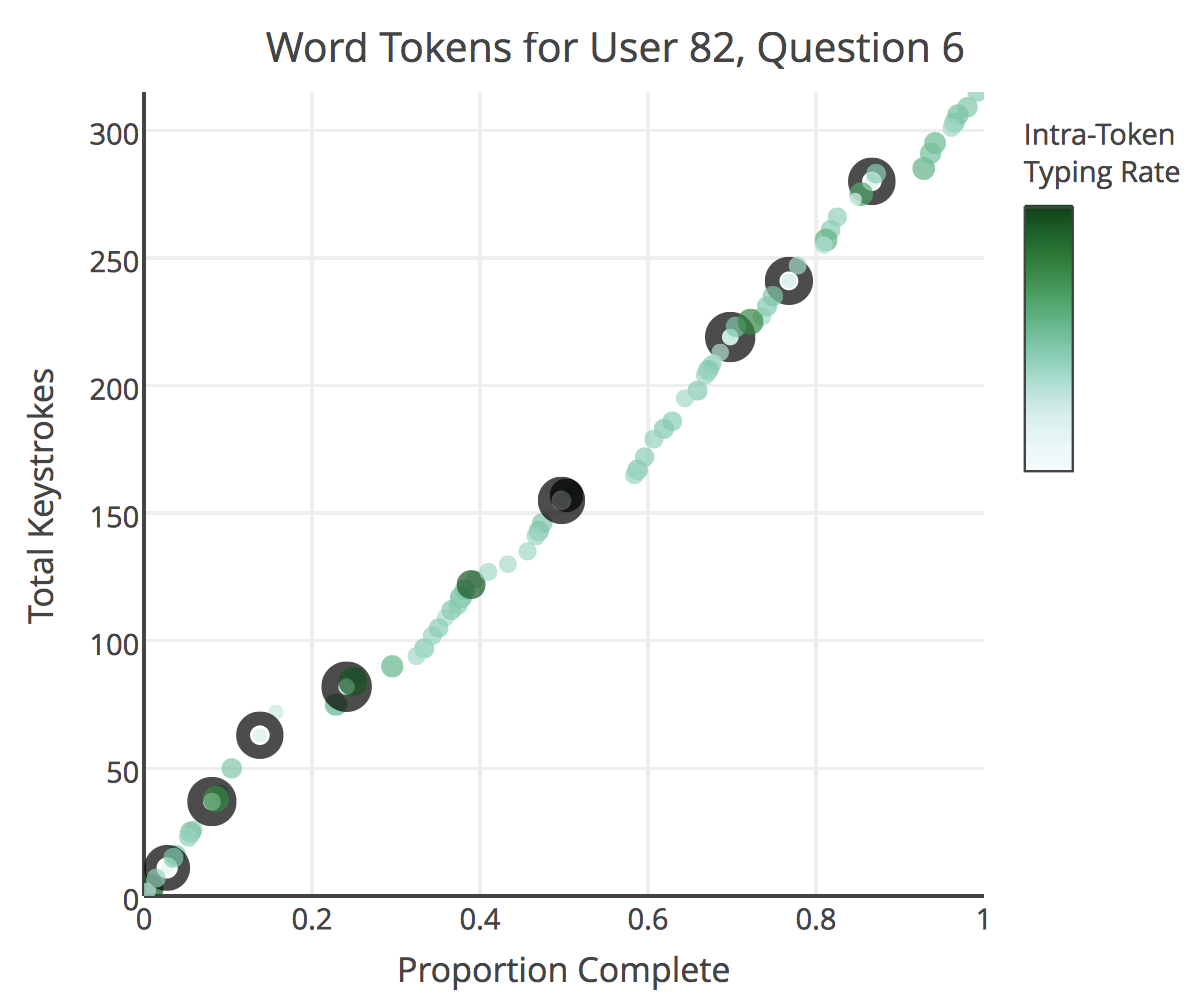}}
    \caption{Heavy revising causes changes in typing dynamics}
    \label{fig:revisingtypist}
\end{figure}

Fig. \ref{fig:revisingtypist} is a non-native English speaker, and looks very different than the previous plots. This typist makes a number of large revisions, as seen by the thick point borders. Likely this is due to less familiarity with English, and hence more spelling or word-choice errors. After the revision, there is often a pause or a change in typing rate. It seems that for this non-native speaker, these errors create greater confusion, and require time to readjust afterwards.

\begin{figure}
    \centering
    \frame{\includegraphics[width=.7\columnwidth]{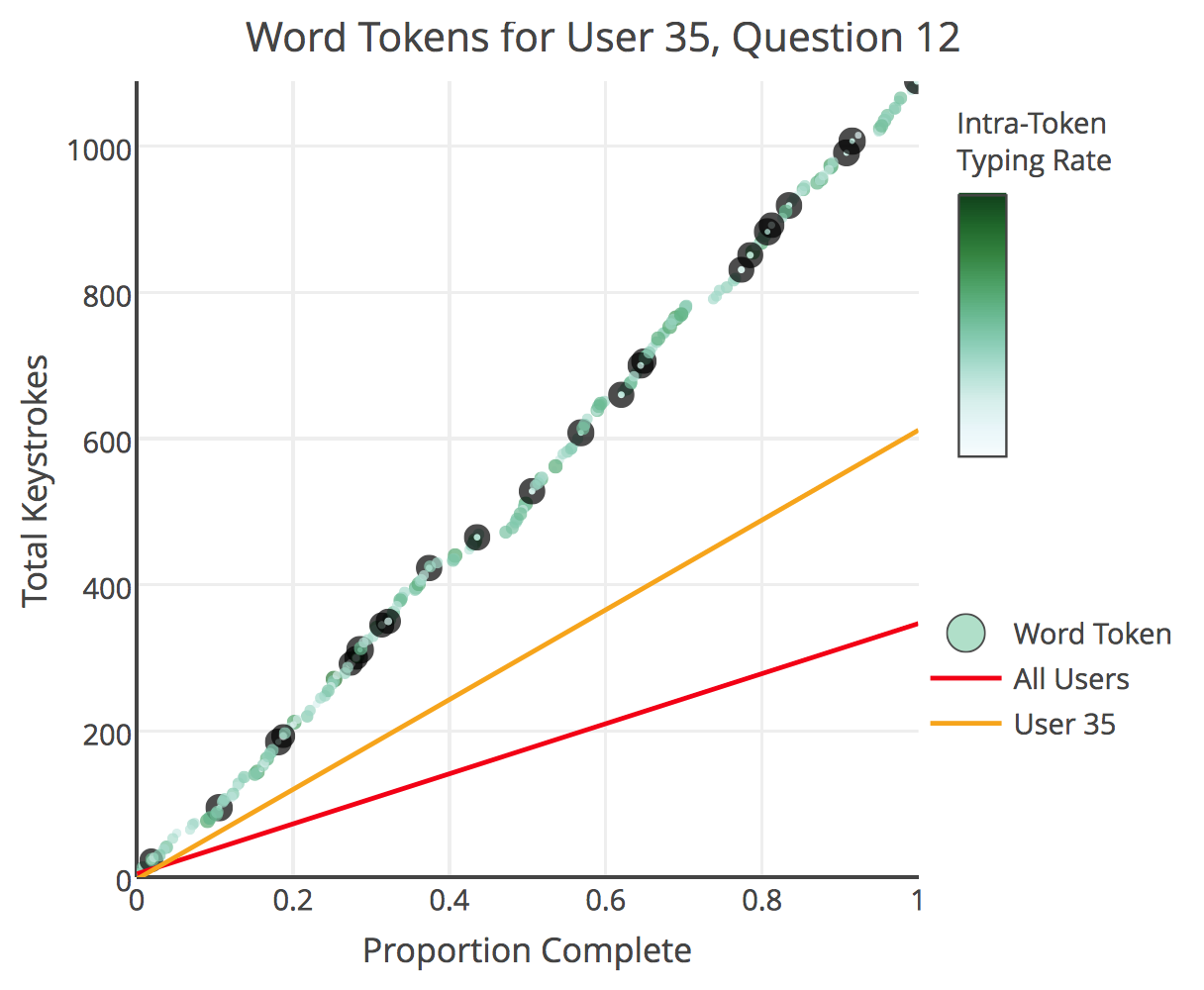}}
    \caption{Faster than average typists that stays consistent}
    \label{fig:fastavg}
\end{figure}

Fig. \ref{fig:fastavg} is a faster-than-average typist, as illustrated by the steeper slope of the user trend line (orange), as compared to the overall trend line (red). Further, this particular session is fast-paced, even for this typist. Unlike Fig. \ref{fig:revisingtypist}, where revisions are a catalyst for a change in typing rate, this more proficient typist is not as affected by revising. Rather, the typist maintains a steady rate, despite frequent revisions. This further sheds light on the importance of comparison trend lines, as they allow us to infer more about the typist than could be inferred from typing session data in isolation.

\begin{figure*}[t]
    \centering
    
    \subfigure[Verbs only \label{fig:subverbs}]{\includegraphics[width=.7\columnwidth]{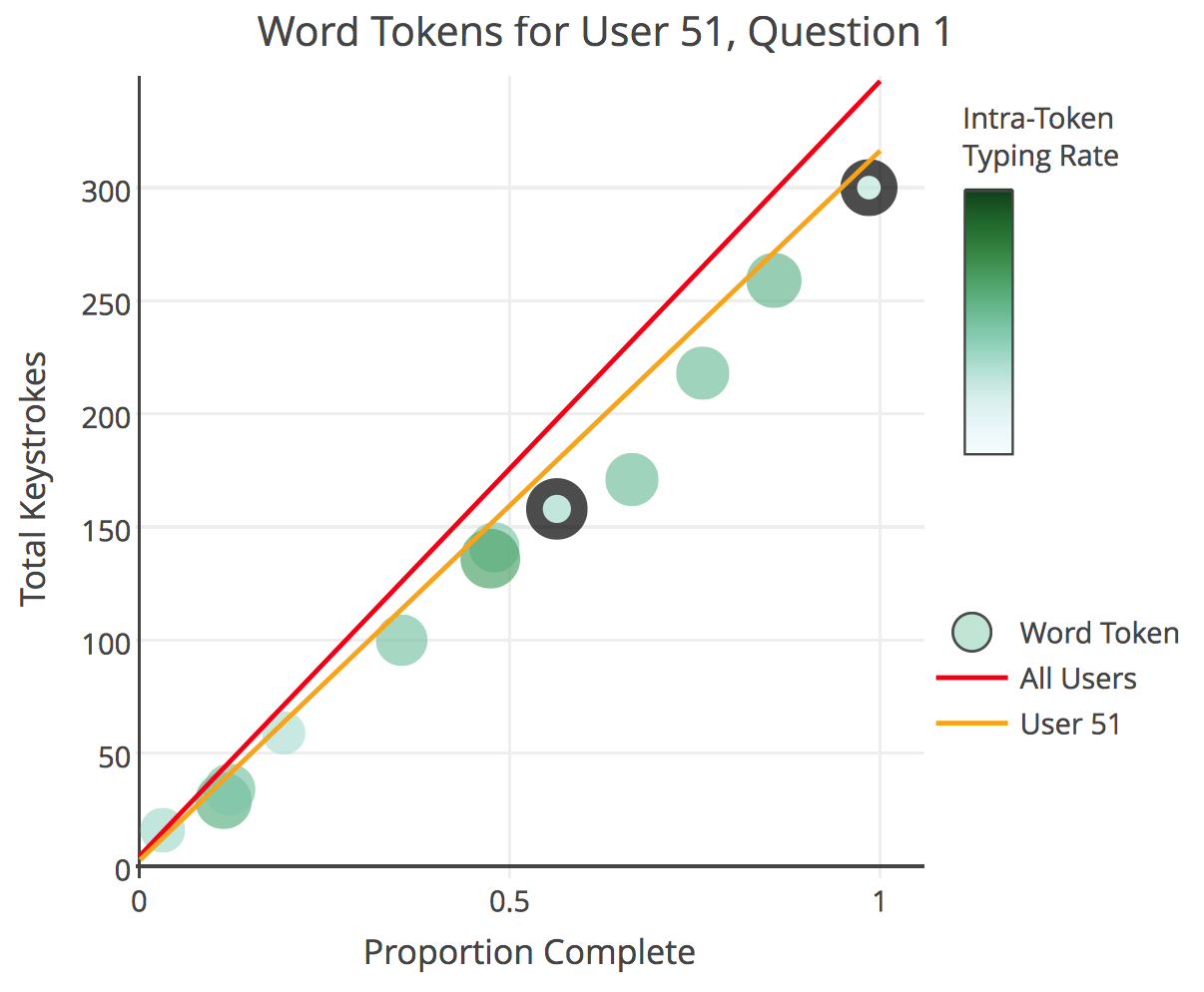}} 
    \subfigure[Nouns only \label{fig:subnouns}]{\includegraphics[width=.7\columnwidth]{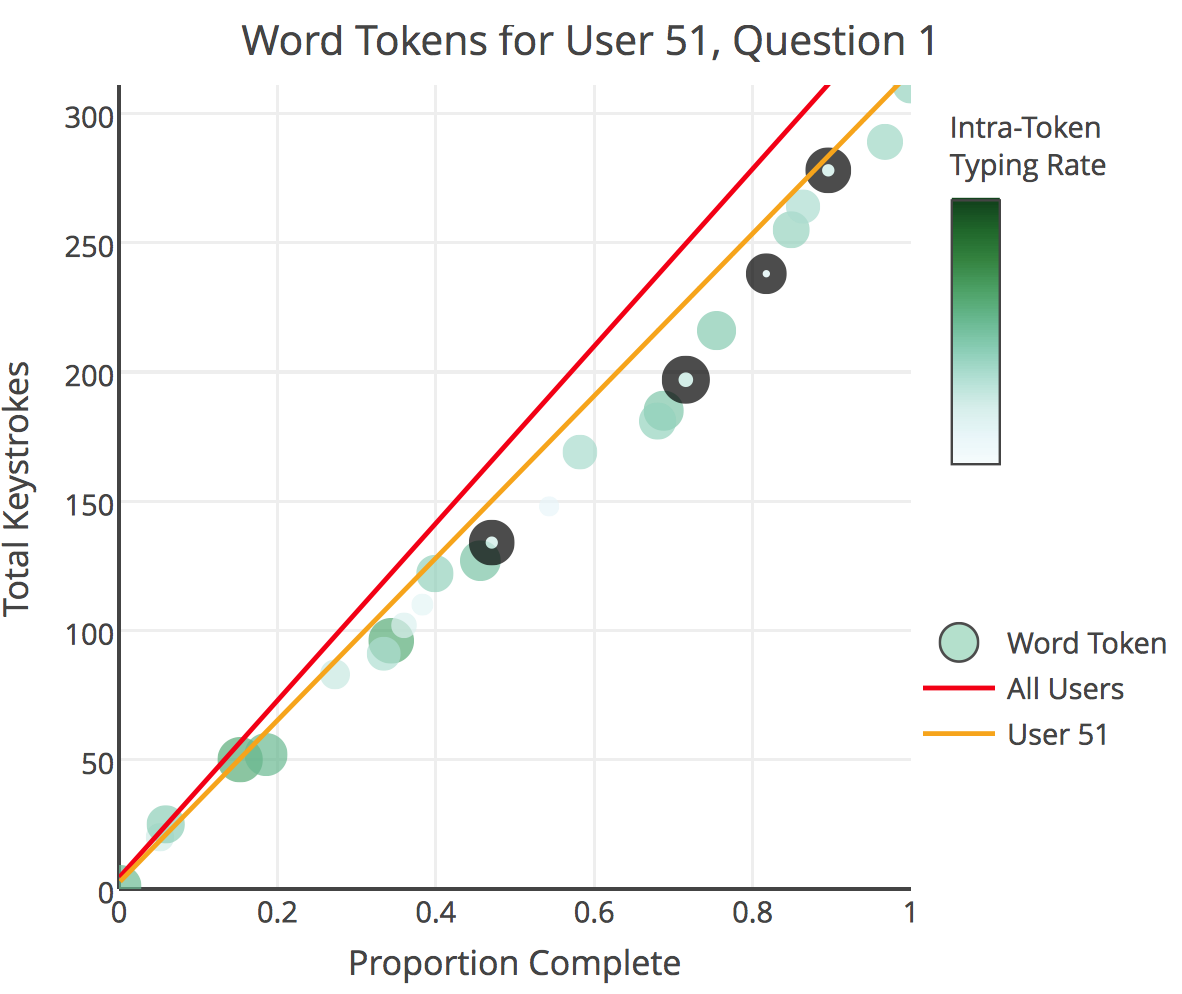}}
    \caption{Comparing typing rates for different parts-of-speech} \label{fig:verbnoun}
\end{figure*}

\begin{figure*}[t]
    \centering
    
    \subfigure[Function words only \label{fig:subfunction}]{\includegraphics[width=.7\columnwidth]{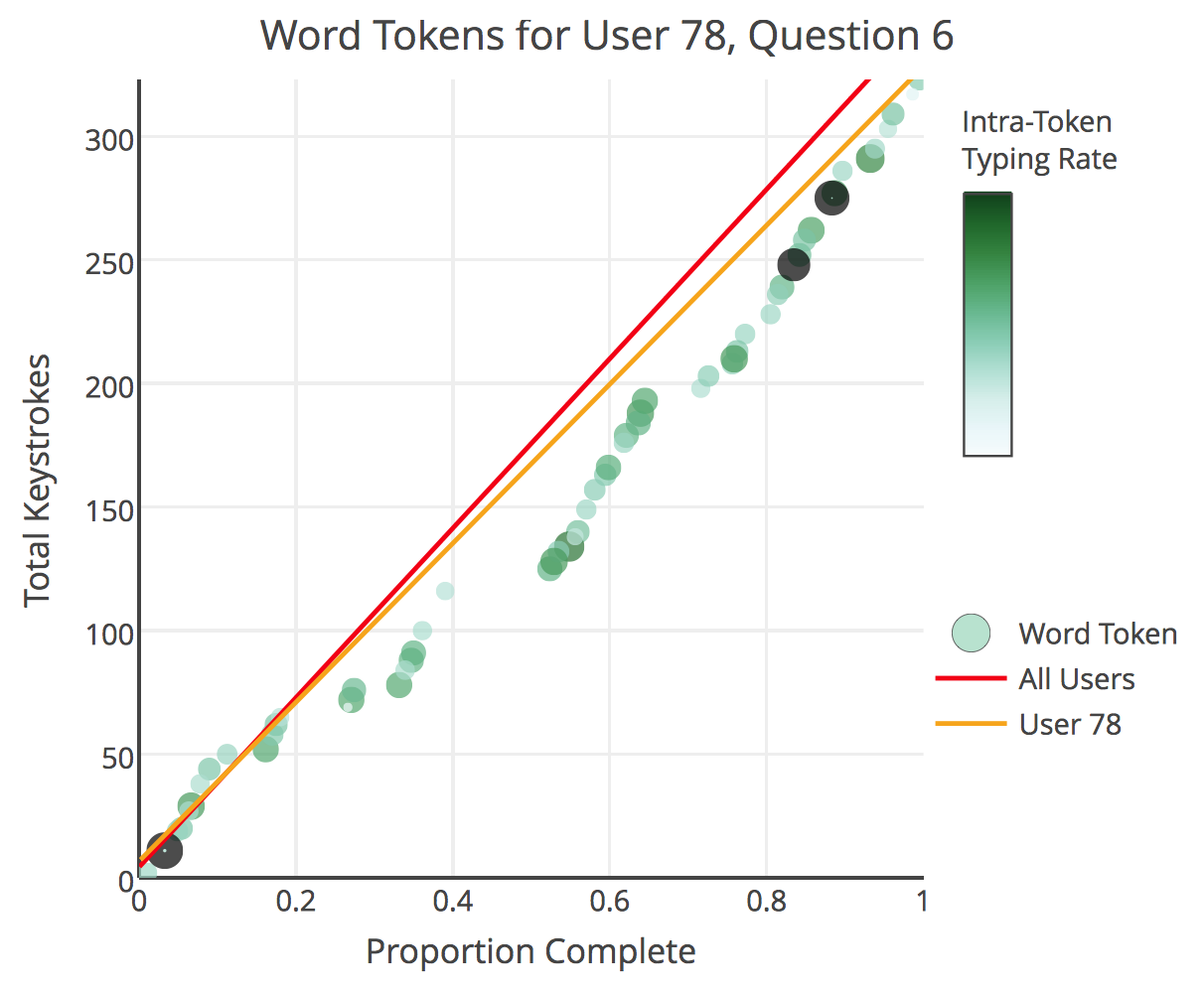}}
    \subfigure[Content words only \label{fig:subcontent}]{\includegraphics[width=.7\columnwidth]{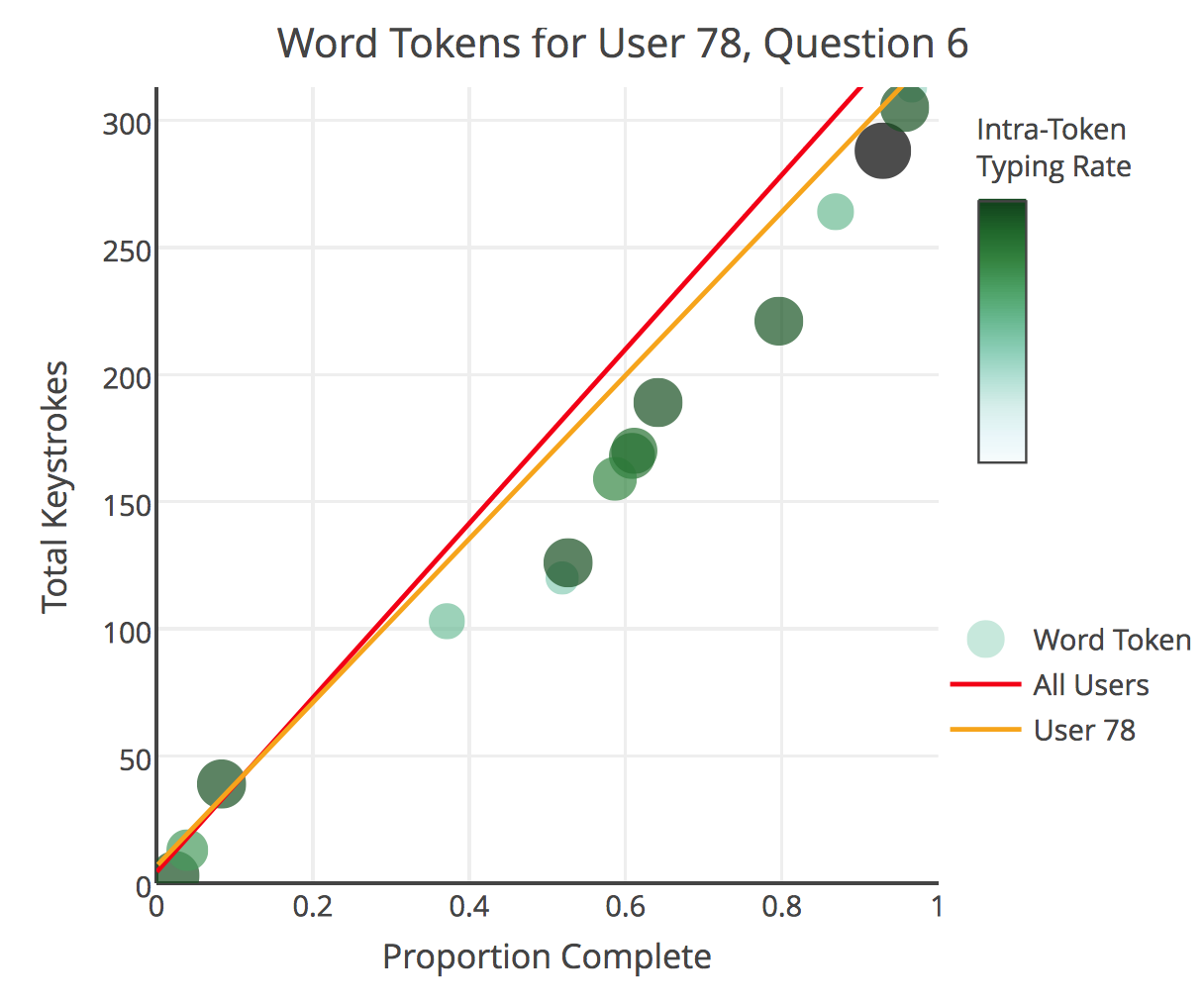}}
    \caption{Comparing typing rates for different semantic categories} \label{fig:functioncontent}
\end{figure*}

In addition to comparing typing rates, being able to filter for only certain types of word tokens may also shed light on interesting trends and differences. 

As can be seen in Figs. \ref{fig:subverbs} and \ref{fig:subnouns}, this user types verbs at a faster rate than she types nouns. This could point to the importance of taking into consideration latent linguistic structure when analyzing a typist's performance. By being able to filter, a user of TypeShift can perform a more direct comparison of these linguistic variables.

Similarly, Figs. \ref{fig:subfunction} and \ref{fig:subcontent} demonstrate how differently a user types function words versus content words. A comparison of these visualizations could be an indicator of distinct cognitive processes driving distinct production rates of different types of words.

\begin{figure}
    \centering
    \includegraphics[width=.7\columnwidth]{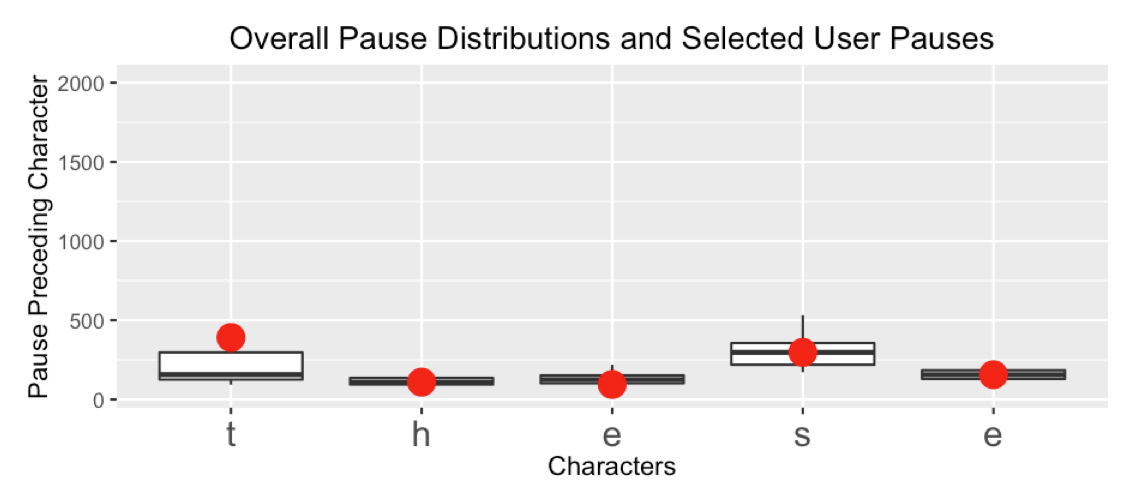}
    \caption{Pre-character pauses for each keystroke in the token ``these''}
    \label{fig:character}
\end{figure}

Finally, as seen in Fig. \ref{fig:character}, a user of TypeShift can zoom in to a particular word token. The red points represents the selected typist's pause durations, while the boxplots represent the overall population distribution of pause durations. This is motivated by the hypothesis that typing production is a hierarchical process: the initial pause is dictated by word retrieval difficulty; once the word is retrieved, subsequent pauses are determined by the difficulty of motor execution \citep{yamaguchi2013hierarchy}. We see evidence for this in Fig. \ref{fig:character}, where the subject is a native Russian speaker. As can be seen, the pause before the initial character is of a longer duration, possibly because of more difficulty retrieving a word in a non-native language. However, the subsequent pauses are very near the population medians, since these pauses are determined by motor skills rather than language familiarity.

\section{Conclusion}

The TypeShift user interface aims to provide a dynamic tool to visualize both the continuous and discrete nature of the language production process. Language production is both a stream of flowing words, as well as a series of separate word tokens. By allowing a user to capture both the holistic process as a single linear progression, as well as highlighting individual characteristics of each particular token, the tool can help a user understand both aspects. Further, by providing comparisons between typing sessions, TypeShift can help illustrate where a typing sessions ranks, both within- and across-subjects.

Ultimately, the goal of understanding language production is to understand how the mind categorizes and processes information. By allowing a user to better visualize and compare typing sessions, more rapid progress can be made towards conceptualizing human information transmission.
\clearpage

\bibliography{acl2020.bib}
\bibliographystyle{acl_natbib}

\end{document}